%% file: arxiv.tex
\documentclass[10pt,journal,compsoc]{IEEEtran}

\usepackage{amsmath,amsfonts}
\usepackage{algorithmic}
\usepackage{algorithm}
\usepackage{array}
\usepackage{textcomp}
\usepackage{stfloats}
\usepackage{url}
\usepackage{verbatim}
\usepackage{graphicx}
\usepackage{cite}

\newcommand{\name}[0]{ADA-Track}

\usepackage{booktabs}
\usepackage{blindtext}
\usepackage{soul}
\usepackage{xcolor}
\usepackage{multirow}
\usepackage{caption}
\usepackage{subcaption}
\usepackage{color, colortbl}

\def\secref#1{Section~\ref{#1}}
\def\figref#1{Figure~\ref{#1}}
\def\tabref#1{Table~\ref{#1}}
\def\eqref#1{Equ.~\ref{#1}}

\makeatletter
\DeclareRobustCommand\onedot{\futurelet\@let@token\@onedot}
\def\@onedot{\ifx\@let@token.\else.\null\fi\xspace}

\def\eg{\emph{e.g}\onedot} 
\def\ie{\emph{i.e}\onedot} 
 
\def\etc{\emph{etc}\onedot}

\makeatother

\usepackage{cite}

\ifCLASSINFOpdf
\else
\fi
\usepackage{tikz}

\begin{document}
%
\title{\name{}++: End-to-End Multi-Camera 3D Multi-Object Tracking with Alternating Detection and Association}

\author{Shuxiao~Ding, Lukas~Schneider, Marius~Cordts, Juergen~Gall

\ifCLASSOPTIONpeerreview
\else
\IEEEcompsocitemizethanks{\IEEEcompsocthanksitem S. Ding, L. Schneider and M. Cordts are with Mercedes-Benz AG, Germany. \\
S. Ding and J. Gall are with University of Bonn, Germany. \\
E-mails: \{shuxiao.ding, lukas.schneider, marius.cordts\}@mercedes-benz.com, gall@iai.uni-bonn.de}
\fi

}

\IEEEtitleabstractindextext{%
\input{sec/0_abstract}

}

\maketitle

\IEEEdisplaynontitleabstractindextext

%
\IEEEpeerreviewmaketitle


\newcommand\submittedtext{%
  \footnotesize This work has been submitted to the IEEE for possible publication. Copyright may be transferred without notice, after which this version may no longer be accessible.}

\newcommand\submittednotice{%
\begin{tikzpicture}[remember picture,overlay]
\node[anchor=north,yshift=-12pt] at (current page.north) {\fbox{\parbox{\dimexpr0.65\textwidth-\fboxsep-\fboxrule\relax}{\submittedtext}}};
\end{tikzpicture}%
}

\submittednotice

%
%
%
%

 




\input{sec/1_intro}
\input{sec/2_related_work}
\input{sec/3_approach}
\input{sec/4_experiment}
\input{sec/5_conclusion}

\ifCLASSOPTIONpeerreview
\else

    \ifCLASSOPTIONcompsoc
      \section*{Acknowledgments}
    \else
      \section*{Acknowledgment}
    \fi

    This work is a result of the joint research project STADT:up (Förderkennzeichen 19A22006O). 
    The project is supported by the German Federal Ministry for Economic Affairs and Climate Action (BMWK), based on a decision of the German Bundestag. 
    The author is solely responsible for the content of this publication. 
    Juergen Gall has been supported by the Deutsche Forschungsgemeinschaft (DFG, German Research Foundation) GA 1927/5-2 (FOR 2535 Anticipating Human Behavior) and the ERC Consolidator Grant FORHUE (101044724).

\fi

\ifCLASSOPTIONcaptionsoff
  \newpage
\fi



%



\bibliographystyle{IEEEtran}
\bibliography{IEEEabrv,ref}

%

\begin{IEEEbiography}[{\includegraphics[width=1in,height=1.25in,clip,keepaspectratio]{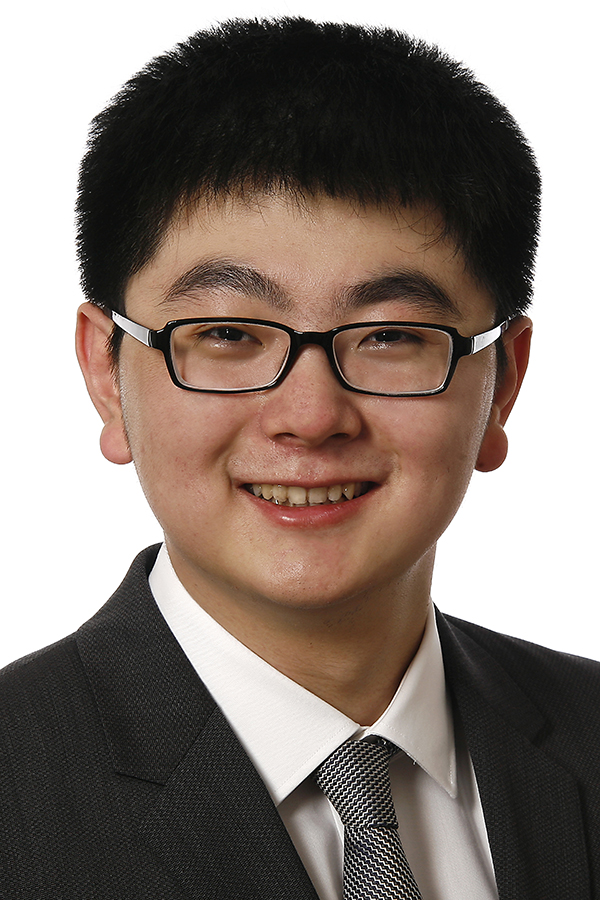}}]{Shuxiao Ding} obtained his B.Sc. in Mechanical Engineering in 2016 and his M.Sc. in Mechatronics and Information Technology in 2019 from Karlsruhe Institute of Technology (KIT). Since 2020, he is a Ph.D. student at the University of Bonn, in cooperation with Mercedes-Benz AG. His research interests include 3D Multi-Object Tracking and Graph Neural Networks for scene understanding in autonomous driving.

\end{IEEEbiography}

\begin{IEEEbiography}[{\includegraphics[width=1in,height=1.25in,clip,keepaspectratio]{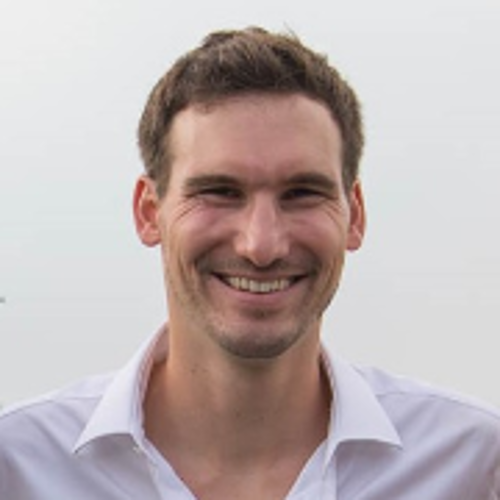}}]{Lukas Schneider} received his Diploma degree in Informatics from Karlsruhe Institute of Technology (KIT) in 2014.
He obtained his Ph.D. degree in computer science from ETH Z\"urich in cooperation with Mercedes-Benz in 2021.
Since 2017, he is working a machine learning engineer at Mercedes-Benz AG in ADAS projects.
His research interests include perception for autonomous driving.

\end{IEEEbiography}

\begin{IEEEbiography}[{\includegraphics[width=1in,height=1.25in,clip,keepaspectratio]{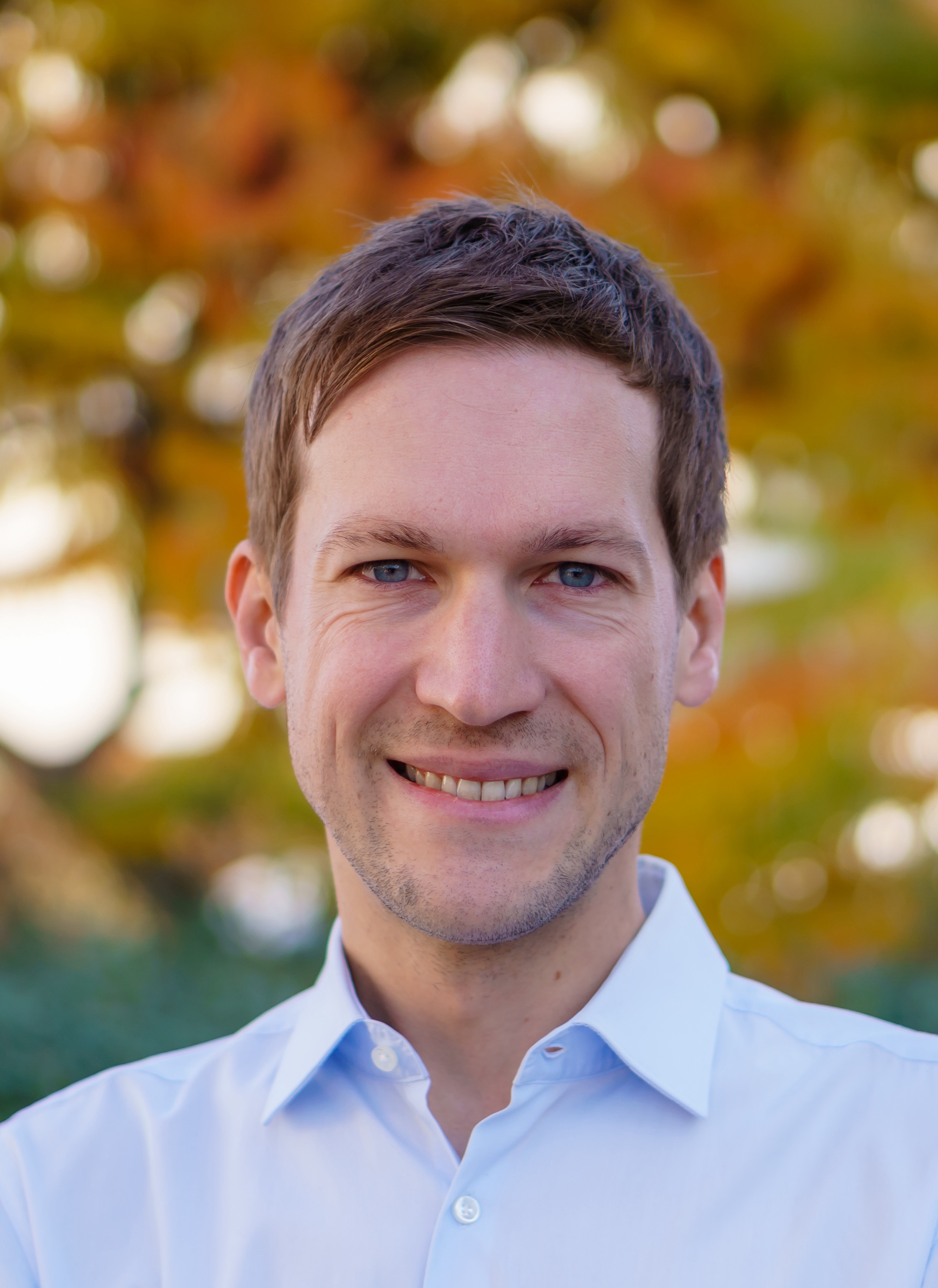}}]{Marius Cordts} obtained his B.Sc. and M.Sc. degrees in Electrical Engineering from the RWTH Aachen University in 2010 and 2013, respectively. In 2017, he received his Ph.D. in computer science from the Technical University of Darmstadt based on his research in cooperation with Mercedes-Benz. During his Ph.D., he published the widely-used Cityscapes dataset. After graduating, he stayed at Mercedes-Benz and worked as a machine learning engineer for various projects around automated driving. Since 2020, he leads a research and development team for computer vision and machine learning.

\end{IEEEbiography}

\begin{IEEEbiography}[{\includegraphics[width=1in,height=1.25in,clip,keepaspectratio]{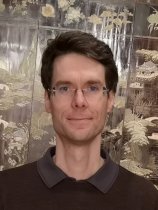}}]{Juergen Gall} obtained his B.Sc. and his Masters degree in mathematics from the University of Wales Swansea (2004) and from the University of Mannheim (2005). In 2009, he obtained a Ph.D.\ in computer science from the Saarland University and the Max Planck Institut f\"ur Informatik. He was a postdoctoral researcher at the Computer Vision Laboratory, ETH Zurich, from 2009 until 2012 and senior research scientist at the Max Planck Institute for Intelligent Systems in T\"ubingen from 2012 until 2013. Since 2013, he is professor at the University of Bonn and PI of the Lamarr Institute for Machine Learning and Artificial Intelligence.
\end{IEEEbiography}
    
    
    




\end{document}

%% file: sec/0_abstract.tex
\begin{abstract}

Many query-based approaches for 3D Multi-Object Tracking (MOT) adopt the tracking-by-attention paradigm, utilizing track queries for identity-consistent detection and object queries for identity-agnostic track spawning.
Tracking-by-attention, however, entangles detection and tracking queries in one embedding for both the detection and tracking task, which is sub-optimal. 
Other approaches resemble the tracking-by-detection paradigm and detect objects using decoupled track and detection queries followed by a subsequent association. 
These methods, however, do not leverage synergies between the detection and association task.
Combining the strengths of both paradigms, we introduce \name{}++, a novel end-to-end framework for 3D MOT from multi-view cameras. 
We introduce a learnable data association module based on edge-augmented cross-attention, leveraging appearance and geometric features. 
We also propose an auxiliary token in this attention-based association module, which helps mitigate disproportionately high attention to incorrect association targets caused by attention normalization.
Furthermore, we integrate this association module into the decoder layer of a DETR-based 3D detector, enabling simultaneous DETR-like query-to-image cross-attention for detection and query-to-query cross-attention for data association. 
By stacking these decoder layers, queries are refined for the detection and association task alternately, effectively harnessing the task dependencies.
We evaluate our method on the nuScenes dataset and demonstrate the advantage of our approach compared to the two previous paradigms.    

\end{abstract}

\begin{IEEEkeywords}
3D Multi-object tracking, multi-camera 3D perception, autonomous driving
\end{IEEEkeywords}

%% file: sec/1_intro.tex
\ifCLASSOPTIONcompsoc
\IEEEraisesectionheading{\section{Introduction}\label{sec:intro}}
\else
\section{Introduction}
\label{sec:intro}
\fi

\IEEEPARstart{A}{ccurate} and consistent 3D Multi-Object Tracking (MOT) is critical for ensuring the reliability and safety of autonomous driving.
Recently, vision-centric perception solely relying on multi-view cameras has garnered significant attention in the autonomous driving community, thanks to lower cost of sensors and the advancements of transformers for computer vision.
Within this domain, two predominant approaches have emerged:
one transforms multi-view features into an intermediate dense Bird's-Eye View (BEV) representation~\cite{li2022bevformer,huang2021bevdet,zhang2022beverse,hu2021fiery,li2023powerbev},
while the other leverages object queries~\cite{carion2020end} that directly interact with the multi-view images~\cite{wang2022detr3d,liu2022petr,doll2022spatialdetr,Wang2023streampetr} to construct an object-centric representation.
Owing to the advantages in modeling object motion, the latter has been extended to query-based MOT in many works~\cite{meinhardt2022trackformer,zeng2022motr,sun2020transtrack,xu2022transcenter,zhao2022tracking}. 

\begin{figure}[t!]
    \centering
    
    \begin{subfigure}{\linewidth}
    \centering
        \includegraphics[width=0.9\textwidth]{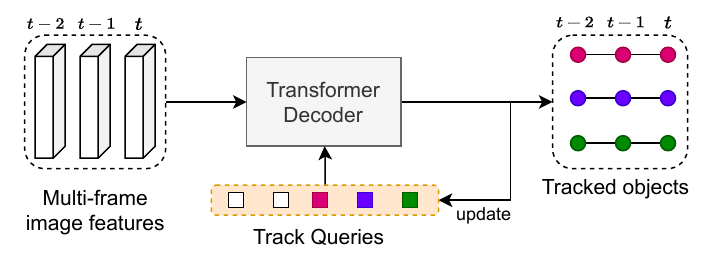}
        \caption{\textbf{Tracking-by-attention} (TBA).}
        \label{subfig:tba}
    \end{subfigure}
    \begin{subfigure}{\linewidth}
    \centering
        \includegraphics[width=0.9\textwidth]{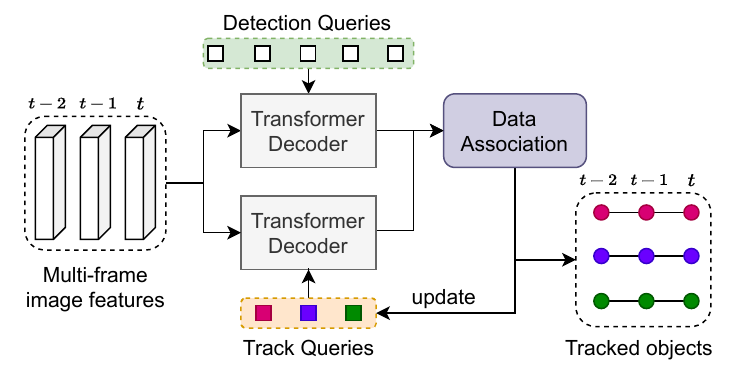}
        \caption{\textbf{Tracking-by-detection} (TBD).}
        \label{subfig:tbd}
    \end{subfigure}
    \begin{subfigure}{\linewidth}
    \centering
        \vspace{2mm}
        \includegraphics[width=0.9\textwidth]{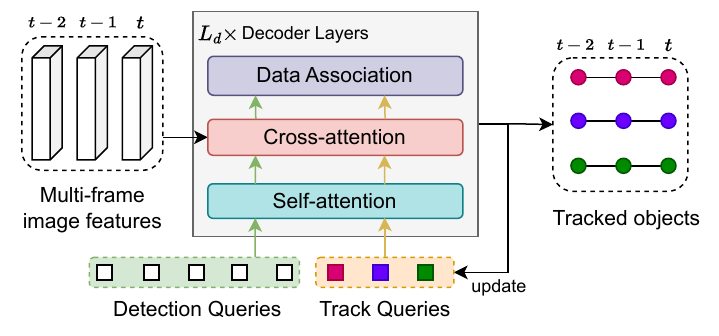}
        \caption{\textbf{Our proposed paradigm.}}
        \label{subfig:ours}
    \end{subfigure}
    \caption{
    Different paradigms of query-based MOT. 
    Our proposed paradigm (\ref{subfig:ours}) leverages the advantages of the coupled architecture of tracking-by-attention (\ref{subfig:tba}) and the decoupled task-specific queries of tracking-by-detection (\ref{subfig:tbd}). 
    }
    \label{fig:titel}
\end{figure}

Among the query-based MOT approaches, the majority adopts the \textit{tracking-by-attention} (TBA) paradigm, \eg~\cite{meinhardt2022trackformer,zeng2022motr}.
As illustrated in~\figref{subfig:tba}, TBA utilizes track queries (colored squares) to consistently detect the same identity across frames and introduces object queries (white squares) to initialize tracks for newly appearing objects in each frame.
However, this highly entangled design is sub-optimal for balancing the detection and tracking performance.
First, each track query consisting of a single embedding is tasked with accomplishing both, detection and tracking, while the two tasks share the same network architecture.
Furthermore, track queries for identity-aware tracking and object queries for identity-agnostic detection are also processed by identical network weights.
We argue that such an approach is sub-optimal for extracting task-specific information from the single query representation.
Second, data association is implicitly addressed using self-attention between all queries.
Although it effectively integrates information of query relations into query refinement, a notable drawback arises during inference.
The network only outputs one confidence score for each object, but it is unclear whether it represents a detection or an association confidence. 
This requires sophisticated manually tuned post-processing. 

Other query-based MOT approaches~\cite{sun2020transtrack,li2022time3d,li2023end} use decoupled detection and track queries to solve detection and tracking tasks independently.
Both query types will be associated explicitly in a heuristic or learnable module, as shown in~\figref{subfig:tbd}.
However, this still inherits the decoupled design of the \textit{tracking-by-detection} (TBD) paradigm and struggles to optimize and harmonize both tasks effectively.

In this work, we argue that detection and tracking is a chicken-and-egg problem: accurate detection enables robust initialization and straightforward association to tracks, while well-established tracks incorporate temporal context to mitigate potential detection errors.
Our method elegantly addresses this challenge by leveraging synergies in both tasks while decoupling them.
We propose \name{}, a novel query-based end-to-end multi-camera 3D MOT framework that conducts object detection and explicit association in an alternating manner, as shown in~\figref{subfig:ours}.
We propagate track queries across frames representing a unique object instance, while generating decoupled detection queries that detect all objects in each frame.
Inspired by~\cite{Ding2023MOTFormer}, we propose a learned data association module based on an edge-augmented cross-attention~\cite{Hussain2021EdgeaugmentedGT}.
In this module, edge features between track and detection queries represent association information. These features are incorporated into attention calculations, updated layer-by-layer, and further used to output affinity scores.
Different from~\cite{Ding2023MOTFormer}, we include appearance features in the nodes and geometric features in the edges, resulting in a fully differentiable appearance-geometric reasoning.
We then integrate the learned association module into each transformer decoder layer of a query-based multi-camera 3D detector, \eg DETR3D~\cite{wang2022detr3d}.
In this way, the decoder layer sequentially conducts a query-to-image cross-attention to refine query representations for object detection and a query-to-query cross-attention to refine query and edge representations for data association.
By stacking the decoder layers, iteratively refined query and edge features provide useful information to each other, resulting in a harmonized optimization of the detection and tracking task.

A preliminary version of the proposed \name{} has been published in~\cite{Ding2024ADA}.
While our earlier work confirms the effectiveness of the novel tracking paradigm, 
we observe that the inherent properties of the softmax function in the cross-attention have some limitations for addressing the data association problem.
In \name{}, we treat detections as queries and tracks as keys within the edge-augmented cross-attention. 
We follow standard attention mechanisms and apply softmax normalization across all existing tracks to compute the attention value towards a detection query, which in turn is used for affinity score estimation.
However, such a detection query does not always have a corresponding track as an association target, \eg, due to newly appearing objects or occlusions. 
In these cases, applying softmax normalization gives a relatively high association score to some tracks, whereas the association score would be ideally the same for all tracks. 
This issue results in wrong associations and thus increases identity switches and degrades tracking performance.   
To address this issue, we propose \name{}++, an improved version of \name{}.
\name{}++ introduces an auxiliary token as an additional association target for the detection queries in the edge-augmented cross-attention.
This modification effectively mitigates the undesirable attention values produced by the softmax normalization.
By incorporating the auxiliary token, the softmax normalization allocates some probability mass to this token, allowing the model to ignore irrelevant tracks by downgrading their attention.
During training, we assign the auxiliary token as the association target for detection queries that should not be associated with any tracks. 

In this improved version, we have also refined the original implementation~\cite{Ding2024ADA}, leading to improved performance and robustness of our method.
To ensure the validity of our results, we have rerun all experiments and report the updated evaluation based on the refined implementation. In addition to these changes, we have conducted more comprehensive experiments to further analyze the proposed \name{} and its enhanced version, \name{}++, including a runtime and complexity analysis, reporting results for additional backbones, and providing additional ablation studies.

We evaluate \name{} and \name{}++ on the nuScenes dataset~\cite{Caesar2019nuScenesAM}.
We compare our proposed alternating detection and association paradigm with approaches based on the other two paradigms.
While achieving promising performance, our proposed paradigm can easily be combined with various query-based 3D detectors.
Compared to \name{}, \name{}++ achieves up to $0.7\%P$ and $1.1\%P$ higher AMOTA based on DETR3D and PETR respectively, demonstrating its stronger ability to handle data association.

%% file: sec/2_related_work.tex
\section{Related Work}
\label{sec:rel_work}

\subsection{Multi-camera 3D detection}
Current research on multi-camera 3D object detection falls into two major categories.
The first category transforms multi-view image features into a dense Bird's-Eye View (BEV) representation using CNNs~\cite{philion2020lift,huang2021bevdet,huang2022bevdet4d,li2023bevstereo,li2023bevdepth,yang2023bevheight} or transformers~\cite{li2022bevformer,yang2023bevformer}.
Although it has been demonstrated that temporal fusion in BEV effectively boosts the detection performance~\cite{huang2022bevdet4d,li2022bevformer,park2022time,zong2023temporal} or supports downstream tasks~\cite{hu2021fiery,zhang2022beverse,li2023powerbev}, such a structured representation may struggle to effectively model object motion.
The second category contains works that follow DETR~\cite{carion2020end}, where sparse object queries interact with multi-view images~\cite{wang2022detr3d,liu2022petr,doll2022spatialdetr,jiang2024far3d}.
This sparse query-based representation facilitates effective object-centric temporal fusion by interacting queries with multi-frame sensor data~\cite{lin2022sparse4d} or propagating queries across frames~\cite{Wang2023streampetr,lin2023sparse4d}.
Consequently, end-to-end detection and tracking methods built on top of query-based detectors have emerged as a popular choice.

\subsection{Tracking-by-attention}
Proposed concurrently by TrackFormer~\cite{meinhardt2022trackformer} and MOTR~\cite{zeng2022motr},
tracking-by-attention (TBA) leverages track queries to detect objects and simultaneously maintain their consistent identities across frames.
TBA regards MOT as a multi-frame set prediction problem, which relies on the self-attention for implicit association and a bipartite matching to force identity-consistent target assignment, similar to a learned duplicate removal~\cite{carion2020end,hosang2017learning,Ding2022End2End}.
MUTR3D~\cite{zhang2022mutr3d} extends the tracking-by-attention paradigm to multi-camera 3D MOT based on a DETR3D~\cite{wang2022detr3d} detector, which additionally updates the 3D reference point of each query besides query feature propagation.
STAR-Track~\cite{doll2023star} improves MUTR3D by proposing a Latent Motion Model (LMM) to update the query appearance feature based on geometric motion prediction.
PF-Track~\cite{pang2023standing} proposes a joint tracking and prediction framework, utilizing a memory bank of queries to refine queries and predict future locations over an extended horizon for occlusion handling.
Despite the fully differentiable design, TBA processes the same query representation using shared network weights for both detection and tracking tasks, which inevitably affects the balance of both tasks.
In this work, we address this problem by introducing decoupled task-dependent queries with a differentiable association module, 
while an alternating optimization strongly couples both tasks in a more reasonable manner. 

\subsection{Tracking by detection}
Many tracking-by-detection (TBD) approaches use a standalone algorithm for data association that can be combined with arbitrary object detectors.
SORT~\cite{Bewley2016SimpleOA}, for instance, uses a Kalman Filter as the motion model and associates the objects using Hungarian Matching~\cite{Kuhn1955TheHM}.
Subsequent works~\cite{Wojke2017SimpleOA,Zhang2021ByteTrackMT,Du2022StrongSORTMD,Aharon2022BoTSORTRA,Cao2022ObservationCentricSR} improve SORT and achieve competitive performance in many 2D MOT benchmarks~\cite{Milan2016MOT16AB,Dendorfer2020MOT20AB}.
Similar pipelines for 3D MOT~\cite{Weng20193DMT,chiu2020probabilistic,pang2022simpletrack,chiu2021probabilistic,wang2023camo, li2023poly,wang2021immortal,benbarka2021score,zhang2023bytetrackv2} have also demonstrated promising performance.
Besides model-based approaches, learning-based methods usually formulate the association problem using a graph structure and solve it using GNNs~\cite{braso2020learning,rangesh2021trackmpnn,zaech2022learnable,nguyen2022lmgp,quach2021dyglip,cheng2023rest} or transformers~\cite{chu2023transmot,zhou2022global,Ding2023MOTFormer}.
TBD in a joint detection and tracking framework has also become a popular choice combined with query-based detectors~\cite{carion2020end,wang2022detr3d}.
These approaches usually decouple detection and track queries, process them independently, and associate them based on IoU~\cite{sun2020transtrack}, box center~\cite{xu2022transcenter}, pixel-wise distribution~\cite{zhao2022tracking} or a learned metric~\cite{li2022time3d,li2023end}.
Although these works are end-to-end trainable, the association module is still separated from the upstream detector, and detection and tracking are processed sequentially, limiting the effective utilization of task dependencies.
In our work, we address this problem by stacking detection and association modules in an alternating fashion. 
In doing so, we utilize the synergies between both tasks.

%% file: sec/3_approach.tex
\section{ADA-Track}
\label{sec:approach}

\begin{figure*}[t!]
    \centering
    \includegraphics[width=0.95\linewidth]{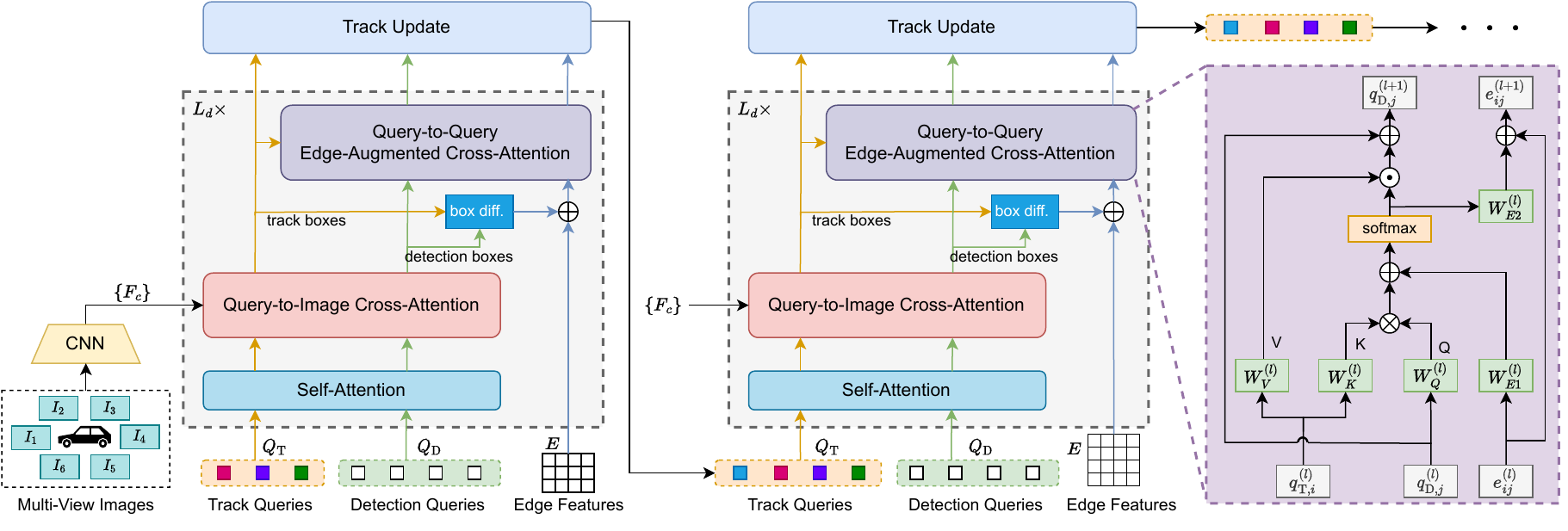}
    \caption{Overview of our \name{} framework. 
    The transformer decoder takes decoupled track and detection queries, zero-initialized edge features, and multi-view image features as input.
    Each decoder layer first refines query features using a self-attention and a query-to-image cross-attention for object detection.
    Then a query-to-query edge-augmented cross-attention is applied to refine detection query and edge features for data association.
    By stacking this decoder layer, query features are updated for both tasks alternately and iteratively. 
    A track update module associates both query sets and produces track queries for the next frame.
    }
    \label{fig:overview}
\end{figure*}

In this section, we introduce the original version of \name{}~\cite{Ding2024ADA}.
An overview is shown in~\figref{fig:overview}.
For each frame $t$, feature maps $F_{c}^t$ are extracted from multi-view images $I_{c}^t$ using a CNN for each camera $c$.
A set of \textit{track queries} $Q_\text{T}^t$ (depicted as colored squares in~\figref{fig:overview}) are propagated from the previous frame in order to consistently detect the same identity across frames.
A fixed number of $N_D$ \textit{detection queries} $Q_\text{D}^t$ (white squares in~\figref{fig:overview}) is randomly initialized and responsible to detect all objects in the current frame.
Following recent works, we assign a 3D reference to each of those queries~\cite{wang2022detr3d,liu2022petr}.
The transformer decoder layer first conducts a \textit{self-attention} between queries and an \textit{image-to-query cross-attention} (\eg DETR3D~\cite{wang2022detr3d} or PETR~\cite{liu2022petr}) to refine queries for the object detection task.
Subsequently, a \textit{query-to-query edge-augmented cross-attention} integrates both query types and edge features and refines them for the data association task.
The overall transformer decoder layer is repeated $L_d$ times to alternately refine query and edge features for both detection and association task.
Finally, a track update module associates the track and detection queries and generates track queries $Q_\text{T}^{t+1}$ for the next frame.

\subsection{Joint detection and association decoder layer}
\label{sec:decoder}
Next, we discuss the decoding process for a single frame $t$ and omit the notation of the frame index $t$ for simplicity.

\subsubsection{Query-to-query self-attention}
Existing approaches with decoupled queries~\cite{sun2020transtrack,li2022time3d,li2023end} process track and detection queries independently and conduct self-attention only within the same query type. 
In contrast, our approach seeks a joint optimization of the representations of both query types.
We concatenate both $Q_\text{T}$ and $Q_\text{D}$ and apply self-attention among all queries regardless of their type.
This self-attention enables detection queries to leverage track queries as prior information, leading to a more targeted interaction with image features in the subsequent layer.

\subsubsection{Image-to-query cross-attention}
Next, both query types $Q_\text{T}$ and $Q_\text{D}$ attend to the multi-view image features $F_{c}$ with cross-attention.
Our approach is compatible with various sparse query-based 3D detectors, thus cross-attention can be implemented in multiple fashions, \eg DETR3D~\cite{wang2022detr3d} or PETR~\cite{liu2022petr}. 
The interaction between queries and images refines the query features to form an object-centric representation.
Subsequently, the network predicts bounding boxes and category confidences using MLPs for each query. This results in track boxes $B_\text{T}^{(l)}$ from track queries and detection boxes $B_\text{D}^{(l)}$ from detection queries, where $l$ denotes the layer index of the decoder layer.
Each box $b_i=[c_i, s_i, \theta_i, v_i] \in \mathbb{R}^9$ is parameterized by 3D box center $c_i \in \mathbb{R}^3$, 3D box size $s_i \in \mathbb{R}^3$, yaw angle $\theta_i \in \mathbb{R}$ and BEV velocity $v_i \in \mathbb{R}^2$.
Both sets of boxes are used to calculate auxiliary box losses as well as to build the position encoding for the edge features that we will discuss next.

\subsubsection{Query-to-query edge-augmented cross-attention}
Our association module requires a differentiable and lightweight architecture that can be integrated into each decoder layer.
To this end, we opt for a learned association module that was recently proposed by 3DMOTFormer~\cite{Ding2023MOTFormer}. 
The module is based on an Edge-Augmented Graph Transformer~\cite{Hussain2021EdgeaugmentedGT} to learn the affinity between tracks and detections.
Different from~\cite{Ding2023MOTFormer}, the query positions change across decoder layers when refining them iteratively in the joint detection and tracking framework.
Therefore, we opt for a fully-connected graph instead of a distance truncated graph to ensure the same graph structure in different layers, enabling the edge features to iterate over layers.

Our learned association leverages both appearance and geometric features.
Formally, for each decoder layer $l$, we use track $Q_\text{T}^{(l)} \in \mathbb{R}^{N_T \times d_k}$ and detection queries $Q_\text{D}^{(l)} \in \mathbb{R}^{N_D \times d_k}$ as node features to provide appearance information obtained from the image-to-query cross-attention,  where $d_k$ is the number of channels. 
An MLP embeds aggregated pair-wise box differences to produce a relative positional encoding $E_\text{pos}^{(l)} \in \mathbb{R}^{N_D \times N_T \times d_k}$ for edge features, \ie $E_\text{pos}^{(l)} = \text{MLP}(B_\text{diff}^{(l)})$, where $B_\text{diff}^{(l)} = \{ b_{\text{diff},ij}^{(l)} \} \in \mathbb{R}^{N_T \times N_D \times 9}$.
These box position features are defined for each pair $\{i,j\}$ of track $b_{\text{T},i}^{(l)} \in \mathbb{R}^9$ and detection boxes $b_{\text{D},j}^{(l)} \in \mathbb{R}^9$ by calculating their absolute difference $b_{\text{diff},ij}^{(l)} = |b_{\text{T},i}^{(l)} - b_{\text{D},j}^{(l)}|$.
The position encoding $E_\text{pos}^{(l)}$ is added to the edge features $E^{(l)} \in \mathbb{R}^{N_D \times N_T \times d_k}$ as part of the input to the edge-augmented cross-attention, \ie $E^{(l)} \xleftarrow{} E_\text{pos}^{(l)} + E^{(l)}$.
As the initial edge features $E^{(0)}$ are zero-initialized for each frame, the input edge features are equal to the edge position encoding for the first layer $l=1$, \ie $E^{(1)} = E_\text{pos}^{(1)}$.

As shown in the right part of~\figref{fig:overview}, we treat track queries $Q_\text{T}^{(l)}$ as source set (key and value) and detection queries $Q_\text{D}^{(l)}$ as target set (query) in the edge-augmented cross-attention.
The edge-augmented attention
\begin{equation}
    A^{(l)} = \text{softmax} \Bigl( \frac{(Q_\text{D}^{(l)} W_Q^{(l)})(Q_\text{T}^{(l)} W_K^{(l)})^T}{\sqrt{d_k}} + E^{(l)} W_{E1}^{(l)} \Bigl)
\end{equation}
takes both dot-product and edge features into consideration, where $\{ W_Q^{(l)}$, $W_K^{(l)} \} \in \mathbb{R}^{d_k \times d_k}$ and $W_{E1}^{(l)} \in \mathbb{R}^{d_k \times 1}$ are learnable weights. 
The feature representation of the targets, \ie the detection queries, as well as the edge features are updated using the attention $A^{(l)} \in \mathbb{R}^{N_D \times N_T \times 1}$, \ie
\begin{equation}
    Q_\text{D}^{(l+1)} = Q_\text{D}^{(l)} + \hat{A}^{(l)}(Q_\text{T}^{(l)} W_V^{(l)}), \:\: E^{(l+1)} = E^{(l)} + A^{(l)} W_{E2}^{(l)},
\end{equation}
where $W_V^{(l)} \in \mathbb{R}^{d_k \times d_k}$ and $W_{E2}^{(l)} \in \mathbb{R}^{1 \times d_k}$ are learnable weights and $\hat{A}^{(l)} \in \mathbb{R}^{N_D \times N_T}$ is $A^{(l)}$ with squeezed third dimension.
The update of $Q_\text{D}^{(l)}$ enables a feature integration of tracking and association information,
resulting in a better query-to-image interaction for the next layer $l+1$.

As there are no existing tracks in the first frame, we skip the edge-augmented cross-attention for $t=1$ and hence the decoder layer becomes identical to the object detector, \eg DETR3D~\cite{wang2022detr3d} or PETR~\cite{liu2022petr}.

\subsection{Track Update}
\label{sec:track_update}
After all $L_d$ layers of the decoder, we obtain the final track and detection queries $Q_T^{(L_d)}$ and $Q_E^{(L_d)}$ as well as edge features $E^{(L_d)}$. 
The track update module associates both query types and propagates feature embeddings of track queries $Q_T^{t+1}$ and their corresponding reference points $C_T^{t+1}$ to $t+1$.

\subsubsection{Data association}
We use different association schemes for training and inference. 
During training, we first match tracks and detections to ground-truth objects (details in~\secref{sec:train}). 
If a track and a detection query are matched to the same ground-truth identity, both queries are considered as a matched pair. 
All track queries that are unmatched with a ground-truth are terminated, while all detection queries that are matched to the newly appearing ground-truth objects spawn a new track.
During inference, we apply an MLP on the final edge features $E^{(L_d)}$ to estimate the affinity scores $S$ between all track-detection pairs, \ie $S=\text{MLP}(E^{(L_d)}) \in \mathbb{R}^{N_D \times N_T}$.
The score matrix $S$ is activated by a sigmoid function and used as matching costs of a Hungarian Algorithm~\cite{Kuhn1955TheHM} (HM), where the element with a pair is rejected in HM if its activated affinity score is lower than a threshold $\tau_S=0.3$.
This produces a one-to-one matching, which produces matched track-detection pairs, unmatched tracks, and unmatched detections.
We heuristically keep the unmatched tracks for $T_d=5$ frames and mark them as temporally inactive tracks before termination.
An unmatched detection box initializes a new track if its confidence is higher than a threshold $\tau_\text{new}=0.4$, following the score threshold for spawning tracks from object queries in~\cite{zhang2022mutr3d,pang2023standing}.

\begin{figure}[!t]
    \centering
    \includegraphics[width=\linewidth]{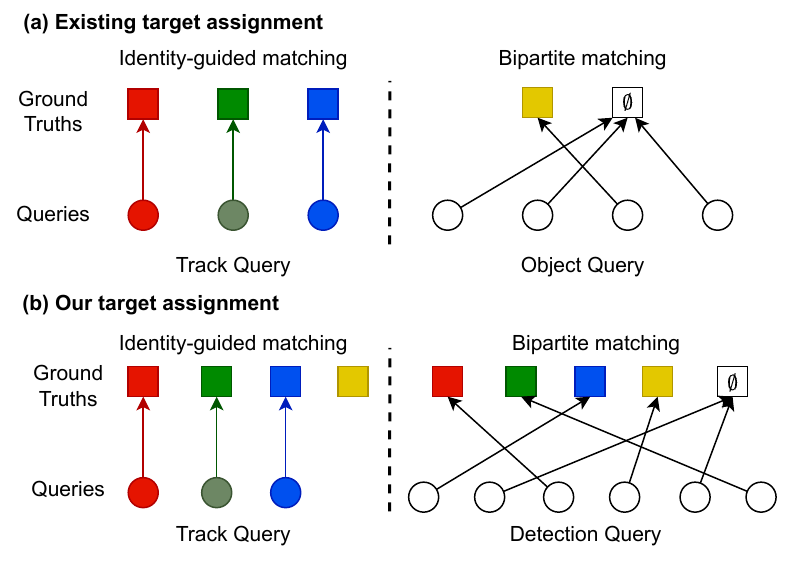}
    \caption{
    Target assignments:
    Tracking-by-attention (a) applies identity-guided matching for track queries and then matches detection queries to remaining ground truths using the Hungarian Algorithm. 
    Our method (b) employs the same matching rules for both query types, but detection queries are matched to all ground-truths.
    }
    \label{fig:target_assign}
\end{figure}

\subsubsection{Feature and box update}
Given a matched pair $\{i,j\}$, the track instance $i$ is assigned with a track query $q_{\text{T},i}^{(L_d)}$ and a predicted track box $b_{\text{T},i}$.
The same is for the detection instance $j$ with detection query $q_{\text{D},j}^{(L_d)}$ and detection box $b_{\text{D},j}$.
Hence, we need to determine the resulting query and box for this associated pair.
Similar to~\cite{Ding2023MOTFormer}, we empirically choose the detection query $q_{\text{D},j}^{(L_d)}$ and detection box $b_{\text{D},j}$ to represent the associated pair $\{i,j\}$, which corresponds to an update of its associated track $i$ at frame $t$: $\hat{q}_{\text{T},i}^{t} = q_{\text{D},j}^{(L_d)}$ and $\hat{b}_{\text{T},i}^{t} = b_{\text{D},j}$.
We will analyse this choice in~\secref{sec:ablation}.
For unmatched detection or track queries, we directly use their respective features and boxes.
Eventually, the track queries $\hat{Q}_{\text{T}}^t = \{\hat{q}_{\text{T},i}^{t}\}$ and their corresponding boxes $\hat{B}_{\text{T}}^t = \{ \hat{b}_{\text{T},i}^{t}\}$ determine the final output of frame $t$.

\subsubsection{Query Propagation}
We directly use the updated query features $\hat{Q}_{\text{T}}^t$ for the next frame, \ie  $Q_{\text{T}}^{t+1} = \hat{Q}_{\text{T}}^t$.
We also propagate their 3D reference point after applying a simple motion update following MUTR3D~\cite{zhang2022mutr3d}.
Concretely, the 3D reference point (box center) $\hat{c}_{\text{T},i}^t$ and the BEV velocity $\hat{v}_{\text{T},i}^t$ are extracted from the box parameter of $\hat{b}_{\text{T},i}^{t}$.
We then predict the reference point for the next frame $t+1$ with a constant velocity assumption: $c_{\text{T},i}^{t+1} = \hat{c}_{\text{T},i}^{t} + \hat{v}_{\text{T},i}^{t} \Delta t$,
where $\Delta t$ is the time difference between both frames.
We transform the predicted reference points $c_{\text{T}}^{t+1}$ to the new vehicle coordinate system with ego-motion compensation.
Together, the track queries combined with the predicted reference points $\{Q_{\text{T}}^{t+1}, C_{\text{T}}^{t+1} \}$ serve as the input for frame $t+1$.

\subsection{Training}
\label{sec:train}

\subsubsection{Target assignment}
\label{sec:tag_ass}
As shown in~\figref{fig:target_assign} (a), tracking-by-attention approaches use identity-guided matching for the track queries, where the bounding boxes of the same identity are always matched to the same track query, while the Hungarian Algorithm~\cite{Kuhn1955TheHM} matches the remaining ground-truth boxes and the object queries, similar to DETR~\cite{carion2020end}.
In this work, track and detection queries detect objects independently and are explicitly associated with each other, which requires isolated matching for both query types.
To achieve that, we use another matching scheme where the matching of both query types is independent, as shown in~\figref{fig:target_assign} (b).
We match all ground-truth identities with detection queries using the Hungarian Algorithm~\cite{Kuhn1955TheHM}, in addition to the same identity-guided matching for track queries.
These matching results define the targets for the bounding box losses.

In addition to the bounding box losses, we apply an explicit loss function for the association module, where we regard it as a binary classification problem.
If a ground-truth identity is matched with both, a track query and a detection query, the edge between these two queries should be classified as positive.
In all other cases, \eg one of both queries is matched with $\emptyset$, or both queries are matched to different ground truths, the association target for this detection-track pair should be negative.

\subsubsection{Loss function}
For the bounding box loss in both query types, we follow~\cite{wang2022detr3d,liu2022petr} and use the Focal Loss~\cite{Lin2017FocalLF} as the classification loss $\mathcal{L}_\text{cls} = \mathcal{L}_\text{FL}$ and $\ell_{1}$-loss as the regression loss $\mathcal{L}_\text{reg} = \mathcal{L}_{\ell_{1}}$.
The track and detection queries contribute to separate loss terms.
This results in $\mathcal{L}_\text{cls,T}$ and $\mathcal{L}_\text{reg,T}$ for the track and $\mathcal{L}_\text{cls,D}$ and $\mathcal{L}_\text{reg,D}$ for the detection part.
For association, we use Focal Loss with $\alpha=0.5$ and $\gamma=1.0$, denoted as $\mathcal{L}_\text{asso} = \mathcal{L}_\text{FL}$.
We include auxiliary losses after each intermediate decoder layer for all the previously mentioned loss terms.

Given a training sequence with $T$ frames, the losses for the detection queries are calculated for all $T$ frames, whereas the losses for track queries and association are calculated from the second frame onwards.
The overall loss for the whole training sequence can be formulated as
\begin{equation}
\begin{aligned}
    \mathcal{L} = & \sum_{t=1}^T (\lambda_\text{cls} \mathcal{L}_\text{cls,D}^t + \lambda_\text{reg} \mathcal{L}_\text{reg,D}^t) + \\
    & \sum_{t=2}^T (\lambda_\text{cls} \mathcal{L}_\text{cls,T}^t + \lambda_\text{reg} \mathcal{L}_\text{reg,T}^t + \lambda_\text{asso} \mathcal{L}_\text{asso}^t).
\end{aligned}
\label{equ:loss}
\end{equation}
We use $\lambda_\text{cls}=2.0,\lambda_\text{reg}=0.25$ following existing works~\cite{wang2022detr3d,liu2022petr} and $\lambda_\text{asso}=10.0$. 
The impact of the loss weight $\lambda_\text{asso}$ is further analyzed in the experiments.

\section{ADA-Track++}
\label{sec:plusplus}
In this section, we present \name{}++, which improves the data association of \name{} by introducing an auxiliary token in the edge-augmented cross-attention.

\subsection{Problem of Attention Normalization in Tracking}
\label{sec:softmax}

\begin{figure}[t!]
    \centering
    
    \begin{subfigure}{\linewidth}
    \centering
        \includegraphics[width=0.85\textwidth]{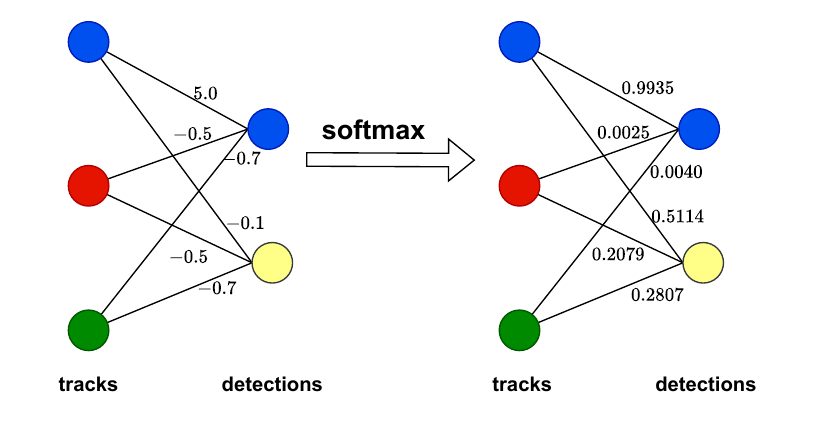}
        \caption{Attentions without auxiliary token.}
        \label{subfig:wo_dummy}
    \end{subfigure}
    \begin{subfigure}{\linewidth}
    \centering
        \includegraphics[width=0.85\textwidth]{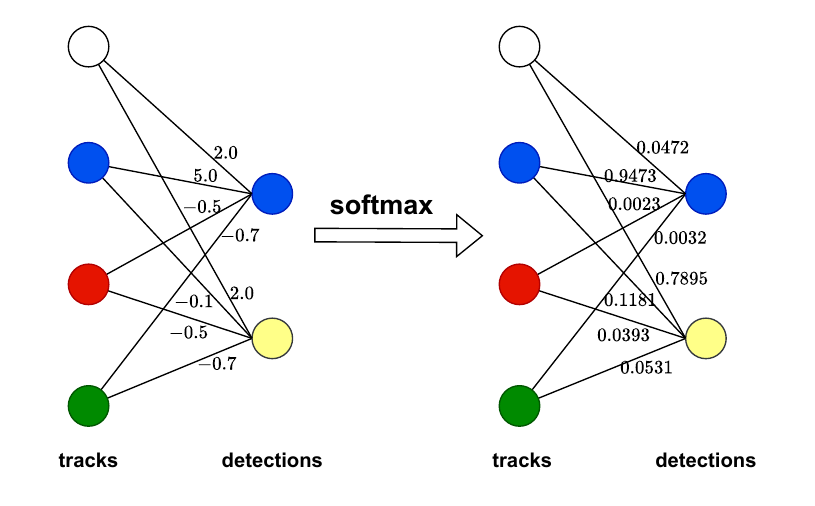}
        \caption{Attentions with auxiliary token (white).}
        \label{subfig:with_dummy}
    \end{subfigure}
    \caption{
    The difference in attention before and after softmax normalization.
    Ground-truth track IDs are represented by unique colors.
    \figref{subfig:wo_dummy}: Although the yellow detection query does not have a corresponding track and its pre-softmax attentions are low everywhere, it still receives relatively high attention to the blue track after softmax normalization, which might disturb the correct association (blue track and blue detection).
    \figref{subfig:with_dummy}: After introducing the auxiliary token, the normalized attention becomes much more reasonable.
    }
    \label{fig:softmax}
\end{figure}

In a transformer’s attention layer, the softmax function is applied to normalize attention scores for each query, converting them into a probability distribution over all keys. 
These normalized attention scores are then used as weights to update the query features through a weighted sum of the values associated with the keys, where the normalization ensures the stability of this feature aggregation.
As depicted in~\figref{fig:overview}, in the edge-augmented cross-attention of \name{}, we regard detections as queries that attend to tracks and aggregate features from them. 
Different from the standard cross-attention, the normalized attention weights are not only used to update the detection features but also to generate the affinity matrix $S$.
In this setup, we aim for each detection query to primarily attend to the track query representing the same instance, while still slightly attending to other track queries to capture scene context or consider alternative association candidates. 
However, challenges arise when a detection query corresponds to a newly appearing object or a false positive. 
In such cases, no track should dominate the attention, but the normalized attention scores will still sum to one, potentially causing the detection query to focus on irrelevant tracks.
To make things worse, even if all pre-softmax attention scores are low for certain detection queries, the highest score will be amplified after softmax (see~\figref{subfig:wo_dummy}).
This results in an unreasonably large probability for assigning the detection to certain tracks, which leads to wrong assignments. 
The normalized attention can further mislead the feature aggregation and affinity estimation and negatively impact the overall performance.

\subsection{Auxiliary token}

To address the issue of normalized attention in the context of learning data association, we propose a simple yet effective solution.
We introduce a randomly initialized auxiliary token $q_{\emptyset} \in \mathbb{R}^{d_k}$ with learnable weights at the first time stamp and propagate it across frames, similar to the track queries. 
In each decoder layer, the auxiliary token participates in the query-to-query self-attention while not attending to the images in the query-to-image cross-attention.
In the edge-augmented cross-attention, we concatenate this auxiliary token to the tracks, \ie, the keys, representing a \textit{None} identity and serving as an additional attention target for the detection queries. 
This produces an attention matrix $A \in \mathbb{R}^{N_D \times (N_T+1)}$, where $N_D$ and $N_T$ denote the number of detection queries and track queries, respectively.
The inclusion of the auxiliary token allows the model to capture detection queries that should not be associated with any existing track queries.
This can be depicted in~\figref{subfig:with_dummy}, where introducing the auxiliary token leads to a desirable attention distribution, which mitigates the risk of irrelevant track queries receiving disproportionate attention, thereby improving the overall data association process.

To ensure that the attention on the auxiliary token is highest when needed, we assign the auxiliary token as the association target in those cases.
Since the association score $S \in \mathbb{R}^{N_D \times (N_T+1)}$ is derived from the attention matrix $A$, the supervision signal on $S$ will be back-propagated to the attention $A$. 
This forces detection queries with no association targets to attend to the auxiliary token.
In \name{}, \ie, without the auxiliary token, the association target for each detection query could either be a one-hot vector, indicating a match with one track, or an all-zero vector, indicating no match with any track. 
With the inclusion of the auxiliary token in \name{}++, all association targets are represented as one-hot vectors. 
Consequently, we can employ categorical cross-entropy as an additional association loss term, which enforces one-hot affinities across tracks for each detection query.
Given $Y_{ij}$ as the association target between the $j$-th detection and the $i$-th track, and $S_{ij}$ representing the affinity before any activation, the additional loss term can be formulated as
\begin{equation}
    \mathcal{L}_\text{CE} = - \sum_{i=1}^{N_D} \sum_{j=1}^{N_T+1} Y_{ij} \log \frac{\exp^{S_{ij}}}{\sum_{k=1}^{N_T+1} \exp^{S_{ik}}}.
\label{equ:loss2}
\end{equation}
Then the overall association loss $\mathcal{L}_\text{asso}$ in~\eqref{equ:loss} becomes a weighted sum of the existing Focal Loss and the cross-entropy loss in~\eqref{equ:loss2}: $\mathcal{L}_\text{asso} = \mathcal{L}_\text{FL} + \lambda_{CE} \mathcal{L}_\text{CE}$, where $\lambda_{CE}=0.1$ is the weight controlling the contribution of the cross-entropy loss term $\mathcal{L}_\text{CE}$ to the overall association loss.

%% file: sec/4_experiment.tex
\section{Experiments}
\label{sec:exp}

\subsection{Implementation Details}
\label{sec:impl}

We evaluate our approach using three image backbones, ResNet-101~\cite{he2016deep}, VoVNet-99~\cite{lee2019energy} and ViT-Large~\cite{dosovitskiy2021an}, and two query-based detectors, DETR3D~\cite{wang2022detr3d} and PETR~\cite{liu2022petr}.
The detection head of all experiments consists of 6 transformer decoder layers.
The model is initialized from the corresponding single-frame detector checkpoints pre-trained for 24 epochs.  
We then train the end-to-end tracker for another 24 epochs on sampled mini-sequences that contain $T=3$ frames.
All models are trained using a cosine-annealing schedule with an initial learning rate of $2e^{-4}$ and an AdamW~\cite{loshchilov2018decoupled} optimizer with a weight decay of $1e^{-2}$.
All experiments with ResNet-101 backbone are conducted on four V100 GPUs, while other backbones use eight A100 GPUs.
Each GPU holds one batch element.
All ablation studies are conducted using DETR3D with ResNet-101.

For all DETR3D-based experiments, input images are full-resolution $1600 \times 900$.
All decoder layers have an embedding dimension of 256 and a feed-forward dimension of 512.
The query embeddings and their corresponding query position encodings of the detection queries are randomly initialized, and the initial reference points are estimated from their initial position encoding using a linear layer.
To better compare with other works, we use $N_D=300$ detection queries and freeze the weights of the image backbone and the FPN during training, following STAR-Track~\cite{doll2023star}.

For all PETR-based experiments, image resolutions are $1600 \times 640$.
The embedding dimension is 256 and the feed-forward dimension is 2048.
In contrast to~\cite{wang2022detr3d}, PETR~\cite{liu2022petr} generates query position encodings using the uniformly initialized reference points while initializing detection queries.
In the query propagation phase, we adhere to this setting to update the query position encodings using the updated reference point positions at each timestamp.
We found this design choice to be essential for ensuring the effectiveness of the PETR-based trackers.
Following PF-Track~\cite{pang2023standing}, we use $N_D=500$ detection queries and only freeze the image backbone for all PETR experiments during training.

In the original version~\cite{Ding2024ADA}, weights of the FPN in the pre-trained detector's checkpoint were not correctly loaded when training the end-to-end tracker, which we fixed in this version.
Another change compared to~\cite{Ding2024ADA} is the update of track queries that spawn new tracks. 
Originally~\cite{Ding2024ADA}, the feature representations of these queries remain unchanged to the initialized features,
whereas features of all track queries should be updated after being processed by the transformer decoder.
In this refined version, features of all track queries are updated before being propagated into the next frame.

\subsection{Dataset and Metrics}
\label{sec:exp_setup}

\begin{table*}[t!]
    \centering
    \setlength{\tabcolsep}{4pt}
    \begin{tabular}{c|c|cc|cccccc|ccc}
        \toprule
         Detector & Tracking Method & AMOTA$\uparrow$ & AMOTP$\downarrow$ & Recall$\uparrow$ & MOTA$\uparrow$ & IDS$\downarrow$ & FP $\downarrow$ & FN $\downarrow$ & TP$\uparrow$ & Params & FLOPs & Inf. time\\
         \midrule
         \multirow{3}{*}{DETR3D~\cite{wang2022detr3d}} & TBA-Baseline & 0.341 & 1.420 & 0.477 & 0.310 & \textbf{466} & 14680 & 41240 & 60191 & 59.73 M & 1024 G & 296 ms \\
          & TBD-Baseline & 0.368 & 1.390 & 0.482 & 0.327 & 987 & \textbf{13294} & 40833 & 60077 & 63.72 M & 1025 G & 297 ms \\
           &\name{} & \textbf{0.391} & \textbf{1.364} & \textbf{0.524} & \textbf{0.343} & 842 & 14790 & \textbf{37923} & \textbf{63132} & 63.73 M & 1054 G & 308 ms\\
         \midrule
         \midrule
         \multirow{3}{*}{PETR~\cite{liu2022petr}} & TBA-Baseline & 0.419 & 1.329 & 0.532 & 0.391 & \textbf{231} & \textbf{13996} & 37425 & 64241 & 82.58 M& 2076 G & 443 ms\\
          & TBD-Baseline & 0.473 & 1.246 & 0.580 & 0.423 & 898 & 14684 & 32884 & 68115 & 97.00 M & 2079 G & 445 ms\\
           & \name{} & \textbf{0.493} & \textbf{1.208} & \textbf{0.602} & \textbf{0.450} & 672 & 14647 & \textbf{30422} & \textbf{70803} & 97.22 M & 2140 G & 453 ms\\
         \bottomrule
    \end{tabular}
    \caption{Comparison of \name{} with TBA- and TBD-baselines on the nuScenes validation set, grouped by DETR3D and PETR detectors. The inference time per frame (\textit{Inf. time}) is measured on an RTX 2080ti GPU.
    }
    \label{tab:baseline}
\end{table*} 

\begin{table*}[t!]
    \centering
    \begin{tabular}{ccc|c|cc|cccccc}
        \toprule
         Method & Backbone & Decoder & Paradigm & AMOTA$\uparrow$ & AMOTP$\downarrow$ & Recall$\uparrow$ & MOTA$\uparrow$ & IDS$\downarrow$ & FP $\downarrow$ & FN $\downarrow$ & TP$\uparrow$\\
         \midrule
         MUTR3D~\cite{zhang2022mutr3d} & R101 & DETR3D & TBA & 0.294 & 1.498 & 0.427 & 0.267 & 3822 & -- & -- & -- \\
         MUTR3D$^\mathparagraph$~\cite{zhang2022mutr3d} & R101 & DETR3D & TBA & 0.341 & 1.420 & 0.477 & 0.310 & 466 & 14680 & 41240 & 60191 \\
         DQTrack~\cite{li2023end} & R101 & DETR3D & TBD & 0.367 & 1.351 & -- & -- & 1120 & -- & -- & -- \\
         STAR-Track$^\dagger$~\cite{doll2023star} & R101 & DETR3D & TBA & 0.379 & 1.358 & 0.501 & 0.360 & \textbf{372} & -- & -- & -- \\

         \rowcolor{gray!30} \name{}$^\S$ \cite{Ding2024ADA} & R101 & DETR3D & ADA & 0.378 & 1.391 & 0.507 & 0.343 & 981 & 15443 & 38466 & 62450 \\
         \rowcolor{gray!30} \name{} & R101 & DETR3D & ADA & 0.391 & 1.364 & 0.524 & 0.343 & 842 & \textbf{14790} & 37923 & 63132 \\
         \rowcolor{gray!50} \name{}++ & R101 & DETR3D & ADA & \textbf{0.398} & \textbf{1.336} & \textbf{0.544} & \textbf{0.357} & 774 &  15345 & \textbf{36381} & \textbf{64742} \\
         
         \midrule
         
         \rowcolor{gray!30} \name{} & V2-99 & DETR3D & ADA & 0.423 & \textbf{1.298} & 0.550 & \textbf{0.391} & \textbf{732} & \textbf{12543} & 35869 & 65296 \\
         \rowcolor{gray!50} \name{}++ & V2-99 & DETR3D & ADA & \textbf{0.434} & 1.304 & \textbf{0.556} & 0.386 & 863 & 14723 & \textbf{34529} & \textbf{66505} \\
         \midrule

         \rowcolor{gray!30} \name{} & ViT-L & DETR3D & ADA & 0.512 & 1.190 & \textbf{0.637} & 0.462 & 793 & \textbf{15228} & 29280 & 71824  \\
         \rowcolor{gray!50} \name{}++ & ViT-L & DETR3D & ADA & \textbf{0.522} & \textbf{1.178} & 0.630 & \textbf{0.465} & \textbf{664} & 15444 & \textbf{29275} & \textbf{71958} \\

         \midrule
        
         MUTR3D$^\mathparagraph$~\cite{zhang2022mutr3d} & V2-99 & PETR & TBA & 0.407 & 1.357 & 0.511 & 0.369 & 271 & \textbf{14401} & 39487 & 62139\\
         PF-Track$^\ddagger$~\cite{pang2023standing} & V2-99 & PETR & TBA & 0.479 & 1.227 & 0.590 & 0.435 & \textbf{181} & -- & -- & --\\
         \rowcolor{gray!30} \name{}$^\S$ \cite{Ding2024ADA} & V2-99 & PETR & ADA & 0.479 & 1.246 & 0.602 & 0.430 & 767 & 15385 & 31402 & 69728 \\
         \rowcolor{gray!30} \name{} & V2-99 & PETR & ADA & 0.493 & 1.208 & 0.602 & \textbf{0.450} & 672 & 14647 & \textbf{30422} & \textbf{70803} \\
         \rowcolor{gray!50} \name{}++ & V2-99 & PETR & ADA & \textbf{0.504} & \textbf{1.197} & \textbf{0.608} & 0.445 & 613 & 14839 & 30616 & 70668\\
         \bottomrule
         
    \end{tabular}
    \caption{
    Comparison of \name{} with existing works on the nuScenes validation set.
    $^\mathparagraph$ denotes reproduced version.
    $^\S$ denotes previous version~\cite{Ding2024ADA}.
    $^\dagger$ denotes total training epochs of 48. 
    $^\ddagger$ indicates annotations from future frames are required.
    }
    \label{tab:val}
\end{table*}

\begin{table}[t!]
    \centering
    \setlength{\tabcolsep}{4pt}
    \begin{tabular}{c|ccccccc}
        \toprule
         Method & AMOTA$\uparrow$ & AMOTP$\downarrow$ & Recall$\uparrow$ & MOTA$\uparrow$ & IDS$\downarrow$\\
         \midrule
         DEFT~\cite{chaabane2021deft} & 0.177 & 1.564 & 0.338 & 0.156 & 6901 \\
         QD-3DT~\cite{hu2022monocular} & 0.217 & 1.550 & 0.375 & 0.198 & 6856 \\
         CC-3DT~\cite{fischer2022cc} & 0.410 & 1.274 & 0.538 & 0.357 & 3334 \\
         \midrule
         Time3D~\cite{li2022time3d} & 0.214 & 1.360 & -- & 0.173 & -- \\
         MUTR3D~\cite{zhang2022mutr3d} & 0.270 & 1.494 & 0.411 & 0.245 & 6018 \\
         PF-Track~\cite{pang2023standing} & 0.434 & 1.252 & 0.538 & 0.378 & \textbf{249} \\
         STAR-Track~\cite{doll2023star} & 0.439 & 1.256 & 0.562 & 0.406 & 607 \\
         \rowcolor{gray!30}
         \rowcolor{gray!50}
         \name{}++ & \textbf{0.500} & \textbf{1.144} & \textbf{0.595} & \textbf{0.456} & 658 \\
         \bottomrule
    \end{tabular}
    \caption{Results on the nuScenes test split. The bottom part compares our method with other end-to-end query-based methods that use DETR3D or PETR. The top part shows other methods.}
    \label{tab:test}
\end{table}

We evaluate our approach on the nuScenes~\cite{Caesar2019nuScenesAM} dataset.
NuScenes is a large-scale dataset for autonomous driving with 700, 150, and 150 sequences for training, validation, and testing.
Each sequence is 20 seconds in length.
The sensor equipment contains LiDAR, RADAR, six cameras covering 360° Field of View as well as IMU and GPS.

The primary metrics for 3D MOT on nuScenes are AMOTA (average multi-object tracking accuracy) and AMOTP (average multi-object tracking precision)~\cite{Weng20193DMT}.
AMOTA is an average of the MOTAR over multiple recalls, where MOTAR is the recall-normalized MOTA at the corresponding recall $r$.
The calculation of MOTA takes IDS (identity switch), FP (false positive) and FN (false negative) into consideration.
AMOTP are the averaged position errors of all TPs (true positive) over all recalls.
NuScenes also uses secondary metrics from CLEAR MOT~\cite{Bernardin2008EvaluatingMO}, including MOTA, MOTP, IDS, FP, FN, \etc.
All the values of the CLEAR MOT metrics are reported at the recall $R$, where the highest MOTA is reached, \ie $R = \text{argmax}_r \text{MOTA}_r$.

\subsection{Paradigm Comparison}
\label{sec:paradigm_comp}
\subsubsection{Baselines}
To validate the superiority of our proposed alternating detection and association paradigm, we first compare our method with query-based tracking-by-attention and tracking-by-detection baselines that are also without bells and whistles.
We select MUTR3D~\cite{zhang2022mutr3d} as the tracking-by-attention baseline by reproducing it on DETR3D and PETR.
For the tracking-by-detection baseline, we modify our architecture to follow~\figref{subfig:tbd}.
To that end, we first use standard DETR3D or PETR decoders to process track and detection queries independently. 
Both sets of queries are then fed into the association module by stacking our proposed query-to-query edge-augmented cross-attentions (\secref{sec:decoder}).
This architecture shares the implementation details with \name{} as described in~\secref{sec:exp_setup}, \eg batch size, training epochs, or weight freezing.
We denote both baselines as \textbf{TBA-Baseline} and \textbf{TBD-Baseline}.

\subsubsection{Results}
\tabref{tab:baseline} compares \name{} with TBA- and TBD-Baselines on the nuScenes validation split.
For both detectors, the TBD-Baselines achieve significantly higher AMOTA than the TBA-Baselines.
On top, \name{} outperforms the TBD-Baselines by 2.3\%P AMOTA for DETR3D and 2.0\%P for PETR, respectively, while achieving a considerably higher recall.
Considering the secondary metrics, we observe that the TBA-Baseline produces fewer IDS than the TBD-Baseline and \name{}, \ie, both methods with the explicit association. 
However, TBA-Baseline achieves a low IDS by tracking only the easy cases, resulting in a substantially higher FN and lower MOTA.
Compared to TBD-Baseline, \name{} again improves MOTA by a much higher TP and lower FN.
These observations show the importance of the decoupled queries with explicit association as in the TBD-Baseline and in \name{}, which produces distinguishable query representations for both tasks and thus improves performances.
\name{} further improves over the TBD-Baseline by fully utilizing the task inter-dependency using alternating detection and association.

We compare the number of parameters, FLOPs, and runtime of \name{} with TBA-Baseline and TBD-Baseline in the right part of~\tabref{tab:baseline}. 
Although \name{} and TBD-Baseline require additional edge-augmented cross-attention modules compared to TBA-Baseline, for DETR3D, \name{} only adds about $6.7\%$ parameters, $2.9\%$ FLOPs, and $4.1\%$ inference time, and even less compared to TBD. 
The same tendency can be observed in PETR-based experiments.
Despite this slight increase in complexity, the performance gain of \name{} is much more significant, \eg, $14.6\%$ and $17.7\%$ higher AMOTA than TBA-Baseline with DETR3D and PETR, respectively.

\subsection{Comparison with Existing Works}
We compare \name{} and \name{}++ with existing end-to-end tracking methods in~\tabref{tab:val}.
First of all, our refined implementation (depicted as \name{}) achieved significantly better performance than reported in the preliminary version~\cite{Ding2024ADA} (depicted as \name{}$^\S$) for both DETR3D and PETR, especially higher AMOTA and lower IDS. This is due to the different checkpoints of the pre-trained detectors as described in \secref{sec:impl}.
For the DETR3D-based experiments, MUTR3D~\cite{zhang2022mutr3d} was already used as TBA-Baseline in~\secref{sec:paradigm_comp}. Interestingly, the original work reported a lower performance than our reproduction with fewer training epochs and a fixed backbone during training. 
DQ-Track~\cite{li2023end} also uses decoupled queries and a sophisticated learned association module following TBD.
\name{} outperforms it by 2.4\%P AMOTA, which again highlights the superiority of our alternating detection and association design.
The training of STAR-Track~\cite{doll2023star} requires an initialization on a pre-trained MUTR3D checkpoint, resulting in a total training epoch of 48.
Our \name{} can still outperform it by $1.2\%$P AMOTA and achieves a much higher recall.
Using larger backbones, \ie, VoVNet-99 and ViT-Large, yields considerable AMOTA increases, demonstrating the generalization to different backbones of \name{}.
We further observe an overall performance gain of \name{}++ compared to \name{} for all backbones, highlighting the effectiveness of introducing the auxiliary token in data association.
For the PETR-based methods, PF-Track~\cite{pang2023standing} is a joint tracking and prediction method, utilizing a track extension module to replace low-confidence detections with predicted trajectories when outputting tracking results, which additionally requires supervision from future frames.
Without any supervision signals from the future, \name{} is a pure tracking method but outperforms PF-Track by $1.4\%P$ AMOTA.
\name{}++ improves \name{} by $1.1\%P$ in AMOTA, demonstrating the consistent benefit of the auxiliary token in \name{}++.

We compare \name{}++ with existing methods on the test split in~\tabref{tab:test},
where we train \name{}++ based on PETR~\cite{liu2022petr} with V2-99~\cite{lee2019energy} backbone on both training and validation splits.
Compared to query-based methods using DETR3D or PETR in the second part of~\tabref{tab:test}, \name{} achieves an AMOTA of 0.500 and an AMOTP of 1.144, outperforming the recent state-of-the-art methods STAR-Track~\cite{doll2023star} and PF-Track~\cite{pang2023standing} by 6.1$\%P$ and 6.6$\%P$ AMOTA, respectively.
Compared to other non-query-based methods, \name{} also achieves the best performance, improving CC-3DT~\cite{fischer2022cc} that used a stronger BEVFormer~\cite{li2022bevformer} detector by 9.0$\%P$ AMOTA.

\begin{table}[t!]
    \centering
    \setlength{\tabcolsep}{3pt}
    \begin{tabular}{c|ccccc|cc}
        \toprule
        $T$ & AMOTA$\uparrow$ & AMOTP$\downarrow$ & Recall$\uparrow$ & MOTA$\uparrow$ & IDS$\downarrow$ & Time & Memory \\
        \midrule 
        2 & 0.321 & 1.429 & 0.478 & 0.286 & 1696 & 1.22 s & 4.04 GB \\
        \textbf{3} & 0.391 & 1.364 & \textbf{0.524} & 0.343 & 842 & 1.44 s & 7.07 GB \\
        4 & \textbf{0.394} & \textbf{1.357} & \textbf{0.524} & \textbf{0.358} & \textbf{766} & 2.15 s & 10.08 GB\\
        \bottomrule
    \end{tabular}
    \caption{Ablation study on the training sample length $T$. \textit{Time} indicates the training time for each step. \textit{Memory} indicates the memory demand during training.}
    \label{tab:frames}
\end{table}

\begin{table}[t!]
    \centering
    \begin{tabular}{cc|ccccc}
        \toprule
        box & asso & AMOTA$\uparrow$ & AMOTP$\downarrow$ & Recall$\uparrow$ & MOTA$\uparrow$ & IDS$\downarrow$ \\
        \midrule
        6 & 1 & 0.351 & 1.390 & 0.476 & 0.305 & 1749 \\
        6 & 2 & 0.378 & 1.371 & 0.494 & 0.329 & 1205 \\
        6 & 3 & 0.389 & 1.361 & 0.516 & 0.344 & 1051 \\
        6 & 4 & 0.389 & 1.359 & 0.519 & 0.340 & 888 \\
        6 & 5 & 0.390 & \textbf{1.358} & 0.522 & \textbf{0.344} & \textbf{820} \\
        \midrule
        1 & 6 & 0.337 & 1.422 & 0.489 & 0.305 & 965 \\
        2 & 6 & 0.379 & 1.373 & 0.491 & 0.336 & 942 \\
        3 & 6 & 0.389 & 1.363 & 0.516 & 0.343 & 925 \\
        4 & 6 & 0.390 & 1.365 & 0.509 & 0.343 & 847 \\
        5 & 6 & 0.387 & 1.368 & 0.502 & 0.343 & 868 \\
        \midrule
        6 & 6 & \textbf{0.391} & 1.364 & \textbf{0.524} & 0.343 & 842 \\
        \bottomrule
    \end{tabular}
    \caption{Ablation study on combining bounding box and association outputs from different decoder layers.}
    \label{tab:layer_output}
\end{table}

\subsection{Ablation Study on \name{}}
\label{sec:ablation}

In this section, we report several ablation studies on \name{} with a focus on understanding the proposed alternating detection and association paradigm.

\subsubsection{Number of training frames}
\tabref{tab:frames} shows the impact of varying the training sample length $T$. 
We observe that an increasing sample length increases the AMOTA and many secondary metrics.
A particularly substantial improvement occurs when increasing from a sample length of 2 to 3 frames, resulting in a notable 7.0\%P increase in AMOTA.
The reason is that the autoregressive training scheme becomes active when $T \ge 3$, allowing association results to propagate into subsequent frames.
This enables optimization through gradients across multiple frames to optimize the whole sequence, significantly boosting robustness during inference. 
A further increase in training frames from 3 to 4 yields marginal improvements but a considerable computational and memory overhead during training.
Thus, we use $T=3$ as the default setting.

\subsubsection{Joint optimization}
One of the main contributions of this work is the introduction of an alternating detection and association paradigm, where both tasks iteratively inform each other layer-by-layer for joint optimization.
To assess its effectiveness, we combine the predictions of the bounding boxes and the association scores from different decoder layers.
The first part of~\tabref{tab:layer_output} illustrates using bounding boxes from the last layer alongside association scores from decoder layers 1 to 5.
A notable increase in AMOTA is observed when using association scores from the second instead of the first layer.
Using association scores from higher layers leads to gradually increased AMOTA.
We also observe a considerable reduction in IDS. 
Together, this shows the iterative improvements of the association.
The second part of~\tabref{tab:layer_output} presents results using boxes from various layers combined with association scores from the last layer.
We can observe a similar tendency in AMOTA increase with higher layers as in the first part, where the most significant increase occurs from the first to the second layer.
In addition, using box predictions in later layers results in a notable improvement in AMOTP, confirming the iterative optimization of localization precision.
Overall, combining outputs from distinct layers yields gradual performance increases when outputs from higher layers are used.
This observation shows that the relevance between box and association predictions is not only constrained within the same layer but across different layers, which validates the iterative optimization through stacked layers of both tasks.

\begin{table}[t!]
    \centering
    \begin{tabular}{c|ccccc}
        \toprule
        $w_t$ & AMOTA$\uparrow$ & AMOTP$\downarrow$ & Recall$\uparrow$ & MOTA$\uparrow$ & IDS$\downarrow$ \\
        \midrule 
        \textbf{0.0} & \textbf{0.391} & 1.364 & \textbf{0.524} & 0.343 & 842 \\
        0.3 & 0.387 & 1.363 & 0.520 & 0.347 & 725 \\
        0.5 & 0.384 & 1.365 & 0.503 & 0.341 & 708 \\
        0.7 & 0.387 & \textbf{1.357} & 0.520 & \textbf{0.358} & 793 \\
        1.0 & 0.382 & 1.369 & 0.502 & 0.341 & \textbf{696} \\
        \bottomrule
    \end{tabular}
    \caption{Impact of the feature update weight $w_t$.}
    \label{tab:feat_weight}
\end{table}

\begin{table}[t!]
    \centering
    \setlength{\tabcolsep}{3pt}
    \begin{tabular}{c|c|c|ccc}
        \toprule
        Exp. &use track box  & track as query & AMOTA$\uparrow$ & AMOTP$\downarrow$ & Recall$\uparrow$ \\
        \midrule 
        A & \checkmark & & 0.387 & 1.357 & 0.510 \\
        B & \checkmark & \checkmark & 0.378 & \textbf{1.347} & 0.483 \\
        \midrule
        C & & & \textbf{0.391} & 1.364 & \textbf{0.524} \\
        \bottomrule
    \end{tabular}
    \caption{Ablation study on box update.
    \textit{use track box} denotes that boxes predicted from track queries are used as output. \textit{track as query} denotes that track queries are regarded as queries and detection queries as keys in the edge-augmented cross-attention.}
    \label{tab:box_update}
\end{table}

\subsubsection{Track update}
\noindent\textbf{Feature update weight } 
We analyze the impact of the feature update as discussed in~\secref{sec:track_update}.
Here we generalize the feature update to a weighted average, \ie, $q_{\text{T},i} = w_\text{T} q_{\text{T},i}^{(L_d)} + (1 - w_\text{T}) q_{\text{D},j}^{(L_d)}$, where $w_\text{T}$ is the update rate.
As shown in~\tabref{tab:feat_weight}, substituting track query features with detection query features ($w_\text{T} = 0$) yields the optimal performance, surpassing other values by 0.4$\%P$ to 0.9$\%P$ in AMOTA.
This can be attributed to the fact that detection queries inherently incorporate track query features within their representation through cross-attention.
Thus, an additional merging between detection and track query features is unnecessary.

\noindent\textbf{Box update } 
Both track and detection queries predict bounding boxes, resulting in two candidates for each instance when two queries are associated.
We use the box from the detection side to represent this associated pair and validate this choice in~\tabref{tab:box_update}. 
As shown in the first row (Exp. A), selecting the track box results in a decrease of 0.4\%P AMOTA and $1.4\%$P recall compared to our approach (Exp. C), which shows the better quality of detection boxes.
One could argue that detection queries are further refined in the edge-augmented cross-attention but track queries are not. 
In the second row of~\tabref{tab:box_update}, we additionally reverse the edge-augmented cross-attention by using tracks as queries and detections as keys. 
This leads to an AMOTA decrease of 1.3\%P and a substantial recall decrease of $4.1\%$P.
This finding also verifies the necessity of the decoupled queries for decoupled tasks, where using detection queries to locate objects and associate them to track queries is more effective than relying on track queries to locate objects.

\begin{table}[t!]
    \centering
    \setlength{\tabcolsep}{3pt}
    \begin{tabular}{c|ccccc}
        \toprule
        rel. pos. encoding & AMOTA$\uparrow$ & AMOTP$\downarrow$ & Recall$\uparrow$ & MOTA$\uparrow$ & IDS$\downarrow$ \\
        \midrule 
        Center & 0.388 & 1.365 & 0.504 & \textbf{0.344} & 895 \\
        None & 0.378 & 1.388 & 0.523 & 0.340 & 1123\\
        Appearance & 0.376 & \textbf{1.357} & 0.515 & 0.342 & \textbf{824}\\
        \midrule
        \textbf{Box} & \textbf{0.391} & 1.364 & \textbf{0.524} & 0.343 & 842\\
        \bottomrule
    \end{tabular}

    \caption{Ablation study on choices of the relative positional encoding in the edge-augmented cross-attention. 
    Our choice is \textit{Box} (last row) which uses all the box parameters to build geometric-based relative position encoding.
    \textit{Center} denotes that only box centers are used. 
    \textit{None} denotes no relative position encoding. 
    \textit{Appearance} uses the query feature differences as appearance-based relative position encoding. 
    }
    \label{tab:edge_feat}
\end{table}

\begin{table}[t!]
    \centering
    \setlength{\tabcolsep}{3pt}
    \begin{tabular}{c|ccccc}
        \toprule
        Edge feat. iteration & AMOTA$\uparrow$ & AMOTP$\downarrow$ & Recall$\uparrow$ & MOTA$\uparrow$ & IDS$\downarrow$ \\
        \midrule 
        &  0.376 & 1.376 & 0.492 & 0.334 & 954 \\
        \checkmark & \textbf{0.391} & \textbf{1.364} & \textbf{0.524} & \textbf{0.343} & \textbf{842}\\
        \bottomrule
    \end{tabular}
    \caption{Ablation study on the iterative refinement of edge features over decoder layers.
    }
    \label{tab:edge_iter}
\end{table}

\subsubsection{Edge features}
\noindent\textbf{Appearance and geometry cues for association}
We investigate the role of appearance and geometric cues in the learnable association module based on edge-augmented cross-attention.
As shown in~\tabref{tab:edge_feat}, using only the center position (first row) instead of the complete box parameters (fourth row) leads to a slight decrease in AMOTA, which underscores the significance of leveraging the entire box information for robust data association.
If geometric features are excluded (second row), the zero-initialized edge features are refined exclusively through appearance-based query features layer-by-layer, resulting in a substantial performance drop across all metrics when compared to the use of geometric-based edge features.
Still without geometric features, using relative positional encodings derived from query feature differences (third row) yields a notable AMOTA decrease of 1.5\%P compared to the default setting (last row).
This observation highlights the usage of geometric features in enhancing the model's ability to distinguish between object instances.

\noindent\textbf{Edge feature refinement}
With an iterative edge feature refinement, we observe a notable decrease in AMOTA by 1.5\%P compared to independent edge features within each decoder layer in~\tabref{tab:edge_iter}.
This experiment shows the potential for iterative optimization of data association across decoder layers, aligning with the fundamental design of our architecture.
In addition, since the edge features also participate in the query feature update in the edge-augmented cross-attention, the refinement of the edge features itself also contributes to the iterative refinement of query representations, which again confirms our architecture design.

\begin{table}[t!]
    \centering
    \setlength{\tabcolsep}{2pt}
    \begin{tabular}{c|ccccc}
        \toprule
        Association module & AMOTA$\uparrow$ & AMOTP$\downarrow$ & Recall$\uparrow$ & MOTA$\uparrow$ & IDS$\downarrow$ \\
        \midrule 
        Node Difference & 0.366 & 1.386 & 0.500 & 0.327 & 1201 \\
        Node Concatenation & 0.364 & 1.388 & 0.500 & 0.327 & 1143 \\
        \midrule
        Edge-Aug. Cross Attn. & \textbf{0.391} & \textbf{1.364} & \textbf{0.524} & \textbf{0.343} & \textbf{842}\\
        \bottomrule
    \end{tabular}
    \caption{Ablation study on different association modules.}
    \label{tab:asso_module}
\end{table}

\begin{table}[t!]
    \centering
    \setlength{\tabcolsep}{2pt}
    \resizebox{\columnwidth}{!}{%
    \begin{tabular}{cc|ccccc}
        \toprule
         \multicolumn{2}{c|}{attention} & \multirow{2}{*}{AMOTA$\uparrow$} & \multirow{2}{*}{AMOTP$\downarrow$} & \multirow{2}{*}{Recall$\uparrow$} & \multirow{2}{*}{MOTA$\uparrow$} & \multirow{2}{*}{IDS$\downarrow$} \\
         det $\rightarrow$ track & track $\rightarrow$ det & \\
        \midrule 
         &  & 0.381 & 1.364 & 0.511 & 0.341 & 856 \\
        \checkmark &  & 0.386 & \textbf{1.358} & 0.502 & \textbf{0.349} & \textbf{805} \\
         & \checkmark & 0.388 & 1.366 & \textbf{0.526} & 0.347 & 968 \\
        \midrule
        \checkmark & \checkmark & \textbf{0.391} & 1.364 & 0.524 & 0.343 & 842 \\
        \bottomrule
    \end{tabular}
    }
    \caption{Ablation study on computing attention across different query types in self-attention.}
    \label{tab:block_attn}
\end{table}

\begin{table}[t!]
    \centering
    \begin{tabular}{c|ccccc}
        \toprule
        $\lambda_\text{asso}$ & AMOTA$\uparrow$ & AMOTP$\downarrow$ & Recall$\uparrow$ & MOTA$\uparrow$ & IDS$\downarrow$ \\
        \midrule 
        2 & 0.378 & \textbf{1.369} & 0.510 & 0.346 & 902 \\
        5 & 0.387 & 1.360 & 0.513 & 0.350 & \textbf{829} \\
        \textbf{10} & \textbf{0.391} & 1.364 & 0.524 & 0.343 & 842 \\
        20 & 0.388 & 1.360 & \textbf{0.531} & \textbf{0.351} & 853 \\
        50 & 0.384 & 1.364 & 0.522 & 0.345 & 875 \\
        \bottomrule
    \end{tabular}
    \caption{Ablation study on the association loss weight $\lambda_\text{asso}$.}
    \label{tab:asso_loss}
\end{table}

\subsubsection{Learned association module}
\tabref{tab:asso_module} compares edge-augmented cross-attention with alternative node-only data association modules in \name{}.
We replace the edge-augmented cross-attention with association networks utilizing the difference or concatenation of detection and track query features (node features). 
In both cases, we use an MLP and sigmoid to obtain the association scores $S$ as before. 
Using only the difference or concatenation of node features results in a significant performance drop, highlighting the necessity of using explicit edge features and the effectiveness of edge-augmented cross-attention in data association.

\subsubsection{Masked self-attention}
The self-attention layer in \name{} facilitates temporal modeling between queries. 
As shown in~\tabref{tab:block_attn}, without interaction between track and detection queries (first row), the model achieves only an AMOTA of 0.381.
In contrast, allowing attention in a single direction, either from detection to track queries or vice versa, improves overall performance. 
Notably, enabling attention from track to detection queries has a more significant positive impact. 
When attention is applied in both directions, the AMOTA improves further, underscoring the importance of bidirectional attention in enhancing query representations and overall performance.

\subsubsection{Association loss weight}
We evaluate the weight of the association loss $\lambda_\text{asso}$ in \name{} in~\tabref{tab:asso_loss}.
Using $\lambda_\text{asso}=10$ yields the highest AMOTA of 0.391. 
Lower or higher association loss weights result in performance drops, which can be attributed to the imbalance of the multi-task training.
Therefore, we choose $\lambda_\text{asso}=10$ as default.

\subsection{Ablation Study on \name{}++}
\label{sec:ablation2}
We further provide more ablation studies on the improved version \name{}++ and compare it with \name{}.

\begin{table}[t!]
    \centering
    \setlength{\tabcolsep}{4pt}
    \begin{tabular}{c|c|ccccc}
        \toprule
        & $N_D$ & AMOTA$\uparrow$ & AMOTP$\downarrow$ & Recall$\uparrow$ & MOTA$\uparrow$ & IDS$\downarrow$ \\
        \midrule 
        \multirow{5}{*}{\shortstack{ADA-\\Track}} & 100 & 0.371 & 1.393 & 0.505 & 0.333 & 1062 \\
        &200 & 0.379 & 1.380 & 0.504 & 0.340 & 837 \\
        &\textbf{300} & \textbf{0.391} & 1.364 & 0.524 & 0.343 & 842 \\
        &400 & 0.389 & 1.365 & 0.524 & 0.347 & 889 \\
        &500 & 0.389 & \textbf{1.363} & \textbf{0.529} & \textbf{0.351} & \textbf{779} \\
        \midrule
        \multirow{5}{*}{\shortstack{ADA-\\Track++}} & 100 & 0.372 & 1.381 & 0.508 & 0.330 & 829 \\
        & 200 & 0.388 & 1.351 & 0.515 & 0.343 & 844 \\
        & 300 & 0.398 & 1.336 & \textbf{0.544} & 0.357 & 774 \\
        & \textbf{400} & \textbf{0.400} & \textbf{1.344} & 0.538 & 0.360 & 777 \\
        & 500 & 0.397 & 1.334 & 0.526 & \textbf{0.359} & \textbf{734}\\
        \bottomrule
    \end{tabular}

    \caption{Impact of the number of detection queries $N_D$ on \name{} and \name{}++.}
    \label{tab:num_query}
\end{table}

\begin{table}[t!]
    \centering
    \setlength{\tabcolsep}{4pt}
    \begin{tabular}{c|c|ccccc}
        \toprule
        & $\tau_S$ & AMOTA$\uparrow$ & AMOTP$\downarrow$ & Recall$\uparrow$ & MOTA$\uparrow$ & IDS$\downarrow$ \\
        \midrule 
        \multirow{5}{*}{\shortstack{ADA-\\Track}} & 0.01 & 0.368 & 1.379 & 0.495 & 0.312 & 1286\\
        & 0.1 & 0.385 & 1.365 & 0.523 & 0.336 & 910\\
        & 0.2 & 0.388 & \textbf{1.358} & 0.496 & 0.335 & \textbf{760}\\
        &\textbf{0.3} & \textbf{0.391} & 1.364 & \textbf{0.524} & \textbf{0.343} & 842 \\
        & 0.4 & 0.382 & 1.365 & 0.512 & 0.339 & 1029\\
        \midrule
        \multirow{5}{*}{\shortstack{ADA-\\Track++}} & 0.01 & 0.388 & 1.338 & 0.525 & 0.334 & 808 \\
        & 0.1 & 0.394 & 1.339 & 0.542 & 0.351 & 716 \\
        & 0.2 & 0.397 & 1.337 & 0.530 & 0.354 & \textbf{689} \\
        & \textbf{0.3} & \textbf{0.398} & \textbf{1.336} & \textbf{0.544} & 0.357 & 774 \\
        & \textbf{0.4} & \textbf{0.398} & 1.338 & 0.531 & \textbf{0.369} & 859 \\
        \bottomrule
    \end{tabular}
    \caption{Impact of the affinity threshold $\tau_S$ on \name{} and \name{}++.}
    \label{tab:aff_thresh}
\end{table}

\begin{table}[t!]
    \centering
    \setlength{\tabcolsep}{4pt}
    \begin{tabular}{c|c|ccccc}
        \toprule
        & $T_D$ & AMOTA$\uparrow$ & AMOTP$\downarrow$ & Recall$\uparrow$ & MOTA$\uparrow$ & IDS$\downarrow$ \\
        \midrule 
        \multirow{5}{*}{\shortstack{ADA-\\Track}} & 2 & 0.378 & 1.389 & 0.514 & \textbf{0.343} & 1156 \\
        & 3 & 0.386 & 1.370 & 0.523 & 0.342 & 952 \\
        & 4 & 0.386 & 1.369 & 0.516 & \textbf{0.343} & 873 \\
        & 5 & 0.391 & 1.364 & \textbf{0.524} & \textbf{0.343} & 842 \\
        & 6 & \textbf{0.392} & \textbf{1.355} & 0.517 & \textbf{0.343} & \textbf{803} \\
        \midrule
        \multirow{5}{*}{\shortstack{ADA-\\Track++}} & 2 & 0.391 & 1.358 & 0.533 & 0.355 & 1074 \\
        & 3 & 0.397 & 1.339 & 0.536 & 0.355 & 886 \\
        & 4 & \textbf{0.399} & \textbf{1.336} & 0.517 & \textbf{0.358} & 791 \\
        & 5 & 0.398 & \textbf{1.336} & \textbf{0.544} & 0.357 & 774 \\
        & 6 & 0.398 & 1.337 & 0.540 & 0.357 & \textbf{741} \\
        \bottomrule
    \end{tabular}
    \caption{Impact of the maximum track age $T_D$ on \name{} and \name{}++.}
    \label{tab:miss_tol}
\end{table}

\subsubsection{Number of queries}

We evaluate the impact of varying the number of detection queries $N_D$ on both \name{} and \name{}++, as shown in \tabref{tab:num_query}. 
Reducing the number of detection queries typically leads to missed detections and a decline in detection performance, which subsequently affects the tracking performance in both variants. 
For \name{}, we observe that AMOTA increases with the number of detection queries up to $N_D=300$. 
However, further increasing $N_D$ does not result in additional improvements.
In contrast, \name{}++ reaches its peak AMOTA at $N_D=400$, indicating its improved ability to accurately associate tracks with a larger number of detections. 
This result underscores the effectiveness of the auxiliary token introduced in \name{}++, which reduces the risk of conflicting affinity scores as discussed in~\secref{sec:plusplus}.
Notably, \name{}++ consistently exhibits lower IDS across all $N_D$ values compared to \name{}, indicating its enhanced ability to reduce confusion in data association.

\subsubsection{Affinity threshold}

During inference, our method filters the estimated affinity score $S$ using a threshold $\tau_S$ before applying the Hungarian Matching (HM). 
\tabref{tab:aff_thresh} presents the impact of this threshold on both \name{} and \name{}++. For both variants, the optimal AMOTA is achieved with $\tau_S=0.3$, while any increase or decrease in the threshold leads to a more significant decline in performance of \name{} than \name{}++. 
For instance, at the threshold $\tau_S=0.01$, where the AMOTA of \name{} drops remarkably to 0.368, \name{}++ maintains a higher AMOTA of 0.388 and an IDS of 808. 
This demonstrates that the affinity scores $S$ in \name{}++ are less noisy, largely due to the introduction of the auxiliary token, which effectively mitigates confusion in the attention scores and enhances the robustness.

\subsubsection{Maximum track age}

We investigate the impact of varying the maximum track age $T_d$ on the performance of \name{} and \name{}++. 
As shown in~\tabref{tab:miss_tol}, low values of $T_d$ result in a significant performance decline for \name{}, particularly at $T_d=2$.
In contrast, \name{}++ maintains an AMOTA of 0.391 at $T_d=2$ with only a slight drop of 0.7 $\%$P compared to the default $T_d=5$.
Additionally, the performance of \name{} stabilizes at $T_d=5$, whereas \name{}++ reaches a plateau starting at $T_d=3$. 
This suggests that \name{}++ is more effective at handling missing tracks, because a lower $T_d$ increases the risk of an early track termination and further leads to inadequate handling of occluded objects or false negative detections. 
This finding further confirms the superiority of \name{}++, where the introduction of the auxiliary token effectively mitigates association confusion caused by detection queries without a corresponding track due to missing targets.

\begin{table}[t!]
    \centering
    \setlength{\tabcolsep}{3pt}
    \begin{tabular}{ccc|ccccc}
        \toprule
         $\mathcal{L}_\text{FL}$ & $\mathcal{L}_\text{CE}$ & activation & AMOTA$\uparrow$ & AMOTP$\downarrow$ & Recall$\uparrow$ & MOTA$\uparrow$ & IDS$\downarrow$ \\
        \midrule 
        \checkmark & & sigmoid & 0.390 & 1.345 & 0.532 & 0.350 & \textbf{715} \\
        & \checkmark & softmax & 0.360 & 1.346 & 0.503 & 0.314 & 1042 \\
        \checkmark & \checkmark & softmax & 0.383 & 1.338 & 0.528 & 0.332 & 1180 \\
        \checkmark & \checkmark & sigmoid & \textbf{0.398} & \textbf{1.336} & \textbf{0.544} & \textbf{0.357} & 774 \\
        \bottomrule
    \end{tabular}
    \caption{Ablation study on the association loss and the activation function of the affinity scores $S$ for \name{}++.}
    \label{tab:asso_loss_act}
\end{table}

\begin{table}[t!]
    \centering
    \begin{tabular}{c|ccccc}
        \toprule
        $\lambda_\text{CE}$ & AMOTA$\uparrow$ & AMOTP$\downarrow$ & Recall$\uparrow$ & MOTA$\uparrow$ & IDS$\downarrow$ \\
        \midrule 
        0.02 & 0.388 & 1.344 & 0.533 & 0.351 & 688 \\
        0.05 & 0.387 & 1.349 & 0.526 & 0.346 & 748 \\
        \textbf{0.1} & \textbf{0.398} & \textbf{1.336} & \textbf{0.544} & \textbf{0.357} & 774 \\
        0.2 & 0.394 & 1.351 & 0.528 & 0.349 & \textbf{677}\\
        0.5 & 0.385 & 1.345 & 0.533 & 0.346 & 682 \\
        \bottomrule
    \end{tabular}
    \caption{Ablation study on the categorical cross entropy loss weight $\lambda_\text{CE}$ in $\mathcal{L}_\text{asso}$.}
    \label{tab:asso_loss2}
\end{table}

\begin{table}[t!]
    \centering
    \setlength{\tabcolsep}{4pt}
    \begin{tabular}{c|ccccc}
        \toprule
         & AMOTA$\uparrow$ & AMOTP$\downarrow$ & Recall$\uparrow$ & MOTA$\uparrow$ & IDS$\downarrow$ \\
        \midrule 
        w/o self-attn & 0.395 & 1.340 & 0.517 & 0.349 & \textbf{695} \\
        w/o prop & 0.396 & \textbf{1.356} & 0.539 & \textbf{0.358} & 786\\
        with prop & \textbf{0.398} & 1.336 & \textbf{0.544} & 0.357 & 774 \\
        \bottomrule
    \end{tabular}
    \caption{Ablation study on the auxiliary token in \name{}++. Our choice (last row) is to propagate the auxiliary token over time stamps and let it participate in the query-to-query self-attention. \textit{w/o self-attn} denotes that the auxiliary token is not involved in the self-attention. \textit{w/o prop} denotes that the auxiliary token is initialized every frame. }
    \label{tab:dummy_feat}
\end{table}

\subsubsection{Loss function and activation}

In~\tabref{tab:asso_loss_act}, we evaluate different combinations of loss functions and activation functions on the affinity scores $S$ of \name{}++. 
When using only a Focal Loss with a sigmoid activation (first row), the performance is highly similar to the standard \name{} with an AMOTA of 0.390. 
This highlights the significance of incorporating the categorical cross-entropy loss to enforce a one-hot output for each detection query, leading to a more effective distribution of affinity scores.
Using a softmax activation (third row) causes a notable AMOTA drop of 1.5$\%$P and a substantial increase in IDS. 
Removing the Focal Loss entirely (second row) leads to a further performance drop with an additional AMOTA decrease of 2.3$\%$P. 
We attribute these performance drops to the relation between the activated affinity scores and the downstream one-to-one matching process. 
Specifically, while the activated affinity serves as the matching cost for the Hungarian matching, which relies on bidirectional bipartite pair matching, the softmax activation normalizes the affinity across one direction, leading to a suboptimal matching cost for HM and declining performance.

\tabref{tab:asso_loss2} evaluates the impact of the categorical cross entropy loss weight $\lambda_\text{CE}$ of \name{}++, where $\lambda_\text{CE}=0.1$ leads to the best overall performance, especially AMOTA and Recall.
Therefore we pick $\lambda_\text{CE}=0.1$ as the default value.

\subsubsection{Auxiliary token feature}

In \tabref{tab:dummy_feat}, we evaluate different choices for the feature representation of the auxiliary token. 
Compared to our default setting (last row), initializing the auxiliary token at each frame (second row) results in a slight AMOTA decrease to 0.396. 
Similarly, when the auxiliary token is excluded from the query-to-query self-attention, a slight decrease in AMOTA is also observed. 
These results confirm our design choice, which allows the auxiliary token to interact more extensively with other queries to capture more scene-level information, leading to a more robust feature representation and improved overall performance.

%% file: sec/5_conclusion.tex
\section{Conclusion}
\label{sec:conclusion}
We presented a novel query-based multi-camera 3D multi-object tracking approach.
We observed that decoupling the detection and association tasks while simultaneously leveraging the synergies between these tasks is the key to accomplish high-quality tracking.
In line with this finding, we proposed a paradigm that conducts both, detection and association, in an alternating manner.
In addition, we proposed a learned association module based on edge-augmented cross-attention which can be seamlessly integrated into any query-based decoder.
To further improve the association process, we incorporated an auxiliary token as an additional association target, reducing the risk of confusion caused by softmax attention normalization.
Extensive experiments show the effectiveness of our approach compared to other paradigms, while achieving promising performance on the nuScenes tracking benchmark. 
Moreover, our method is compatible with 3D detectors and can easily incorporate insights from other tracking methods, making it highly flexible and adaptable to different setups.
We hope that our method can inspire future research in the community and serve as an established baseline for 3D multi-object tracking.